\documentclass[nohyperref,dvipsnames]{article}

\usepackage{microtype}
\usepackage{graphicx}
\usepackage{subfigure}
\usepackage{booktabs} 

\usepackage{hyperref}

\usepackage[accepted]{icml2022}

\usepackage{amsmath}
\usepackage{amssymb}
\usepackage{mathtools}
\usepackage{amsthm}
\usepackage[utf8]{inputenc} 
\usepackage[T1]{fontenc}    
\usepackage{hyperref}       
\usepackage{url}            
\usepackage{booktabs}       
\usepackage{amsfonts}       
\usepackage{nicefrac}       
\usepackage{microtype}      
\usepackage{xspace}
\usepackage{amsmath}
\usepackage{multirow}
\usepackage{multicol}
\usepackage{makecell}
\usepackage{enumitem}

\DeclareMathOperator*{\trace}{Tr}
\DeclareMathOperator*{\dfil}{dFIL}
\DeclareMathOperator*{\att}{Att}
\DeclareMathOperator*{\diam}{diam}

\DeclareMathOperator*{\enc}{Enc}
\DeclareMathOperator*{\conv}{Conv}
\DeclareMathOperator*{\divergence}{div}

\usepackage[capitalize,noabbrev]{cleveref}

\theoremstyle{plain}
\newtheorem{theorem}{Theorem}

\newtheorem{corollary}[theorem]{Corollary}
\theoremstyle{definition}

\theoremstyle{remark}

\usepackage[textsize=tiny]{todonotes}

\icmltitlerunning{Bounding the Invertibility of Privacy-preserving Instance Encoding using Fisher Information}

\begin{document}
\twocolumn[
\icmltitle{Bounding the Invertibility of Privacy-preserving Instance Encoding using Fisher Information}



\icmlsetsymbol{equal}{*}

\begin{icmlauthorlist}
\icmlauthor{Kiwan Maeng}{equal,psu}
\icmlauthor{Chuan Guo}{equal,meta}
\icmlauthor{Sanjay Kariyappa}{gt}
\icmlauthor{G. Edward Suh}{meta,cornell}

\end{icmlauthorlist}

\icmlaffiliation{psu}{Pennsylvania State University}
\icmlaffiliation{meta}{Meta AI}
\icmlaffiliation{gt}{Georgia Institute of Technology (work done while at Meta AI)}
\icmlaffiliation{cornell}{Cornell University}

\icmlcorrespondingauthor{Kiwan Maeng}{kvm6242@psu.edu}

\icmlkeywords{Machine Learning, ICML}

\vskip 0.3in
]




\printAffiliationsAndNotice{\icmlEqualContribution} 

\begin{abstract}

%
%


Privacy-preserving instance encoding aims to encode raw data as feature vectors without revealing their privacy-sensitive information. When designed properly, these encodings can be used for downstream ML applications such as training and inference with limited privacy risk. However, the vast majority of existing instance encoding schemes are based on heuristics and their privacy-preserving properties are only validated empirically against a limited set of attacks.
In this paper, we propose a theoretically-principled measure for the privacy of instance encoding based on Fisher information. We show that our privacy measure is intuitive, easily applicable, and can be used to bound the invertibility of encodings both theoretically and empirically. 
\end{abstract}
\section{Introduction}

\begin{figure}[t]
    \centering
    \includegraphics[width=.49\textwidth]{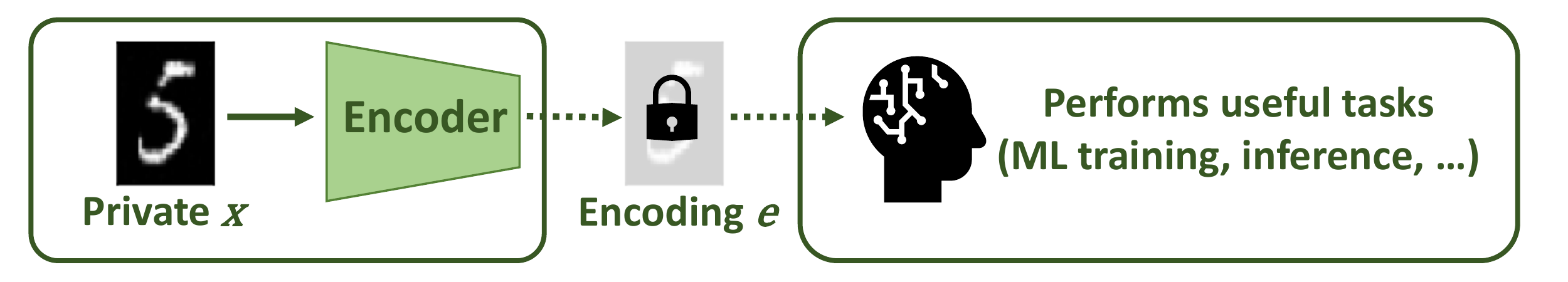}
    \vspace{-4ex}
    \caption{Instance encoding maps an input $x$ to its encoding $e$ that can be used for downstream tasks. The objective is to design encoders such that $e$ reveals very little private information about $x$ while retaining information relevant to the downstream task.
    }
    \vspace{-2ex}
    \label{fig:overview}
\end{figure}

Machine learning (ML) applications often require access to privacy-sensitive data. Training a model to predict a patient's disease with x-ray scans requires access to raw x-ray images that reveal the patient's physiology~\cite{xray}. Next-word prediction for smart keyboards requires the user to input a context string containing potentially sensitive information~\cite{fl_gboard}. To enable ML applications on privacy-sensitive data, \emph{instance encoding} (\citet{instahide_broken}; Figure~\ref{fig:overview}) aims to encode data in a way such that it is possible to run useful ML tasks---such as model training and inference---on the encoded data while the privacy of the raw data is preserved.
The concept of instance encoding is widespread under many different names: learnable encryption~\cite{instahide, neuracrypt, dauntless, imaginary_rotate}, split learning~\cite{split_learning, split_learning2}, split inference~\cite{neurosurgeon, splitnets}, and vertical federated learning (vFL; \citet{fl_survey, sfl, ressfl}) are all collaborative schemes for training or inference that operate on (hopefully) privately-encoded user data.
%

Unfortunately, existing methods for instance encoding largely rely on heuristics rather than rigorous theoretical arguments to justify their privacy-preserving properties.
For example, \citet{instahide, neuracrypt, nopeek, nopeek-infer, ressfl} proposed instance encoding schemes and empirically showed that they are robust against certain input reconstruction attacks. However, these schemes may not be private under more carefully designed attacks; in fact, many encoding schemes that were initially thought to be private have been shown to be vulnerable over time~\cite{instahide_broken, neuracrypt_broken}.


In contrast to prior work, we propose a framework to quantify how easy it is to invert an instance encoding in a theoretically-principled manner using (diagonal) \emph{Fisher information leakage} (dFIL; \citet{fil, fil_guo})---an information-theoretic measure of privacy with similar properties to differential privacy (DP; \citet{dwork2006calibrating, dpbook}). dFIL can be computed for common privacy-enhancing mechanisms and used to lower-bound the expected mean squared error (MSE) of an input reconstruction attack when given the output of the privacy-enhancing mechanism. We apply this reasoning to instance encoding and show that dFIL can serve as a useful measure for encodings' invertibility, by lower-bounding the reconstruction error of an arbitrary attack.
%
%
\emph{To the best of our knowledge, our work is the first to theoretically lower-bound the invertibility of instance encoding for an arbitrary attacker and use it to design practical training/inference systems with high privacy}.

%

%
%

\paragraph{Contributions} Our main contributions are as follows:
\begin{enumerate}[noitemsep, leftmargin=*, topsep=0pt]
    \item We adapt the result of \citet{fil_guo} for instance encoding to show how dFIL can lower bound the MSE of \emph{particular} input reconstruction attacks (\emph{i.e.}, \emph{unbiased} attacks) that aim to reconstruct the raw data given the encoding. We show how popular encoders can be modified minimally for dFIL to be applied (Section \ref{sec:dfil_encoding}).
    \item We extend the result of \citet{fil_guo} and show that dFIL can lower bound the MSE of \emph{any} input reconstruction attack (\emph{e.g.}, strong attacks leveraging knowledge of the input prior; Section \ref{sec:bound_biased}). Our extension involves a novel application of the classical \emph{van Trees inequality}~\cite{vantrees} and connecting it to the problem of \emph{score matching} in distribution estimation.
    \item We evaluate the lower bound using different attacks and encoding functions, and show that dFIL can be used to interpret the privacy of instance encoding both in theory as well as against realistic attacks (Section \ref{sec:eval_bound}).
    \item
    We show how dFIL can be used as a practical privacy metric and guide the design of privacy-enhancing training/inference systems with instance encoding (Section~\ref{sec:split_inference}--\ref{sec:private_training}). We show that it is possible to achieve both high (theoretically-justified) privacy and satisfactory utility.
\end{enumerate}

\section{Motivation and Background}
\label{sec:bg}

\subsection{Instance Encoding}
\label{sec:bg_encoders}

%

Instance encoding is the general concept of encoding raw input $\mathbf{x}$ using an encoding function $\enc$ so that private information contained in $\mathbf{x}$ cannot be inferred from its encoding $\mathbf{e} = \enc(\mathbf{x})$.
%
The principle behind the privacy-preserving property of instance encoding is that the function $\enc$ is hard to invert. However, prior works generally justify this claim of non-invertibility based on heuristics rather than rigorous theoretical analysis~\cite{nopeek, nopeek-infer, ressfl}.
Alternatively, \citet{neuracrypt, dauntless, imaginary_rotate} proposed to use a \emph{secret} encoder network for private training, whose privacy guarantee relies on the secrecy of the encoder network. Such approaches can be vulnerable if the secret  is revealed, which can happen when enough input-encoding pairs are observed~\cite{dauntless, neuracrypt_broken}. 


\paragraph{Attacks against instance encoding}
Given an instance encoder $\enc$, the goal of a reconstruction attack is to recover its input.
Formally, let $\mathbf{e} = \enc(\mathbf{x})$ be the encoding of an input $\mathbf{x}$, and let $\att$ be an attack that aims to reconstruct $\mathbf{x}$ from $\mathbf{e}$: $\hat{\mathbf{x}} = \att(\mathbf{e})$.
Such an attack can be carried out in several ways. If $\enc$ is known, $\hat{\mathbf{x}}$ can be obtained by solving the following optimization~\cite{attacks}:
$\hat{\mathbf{x}} = \underset{\mathbf{x_0}}{\arg\min}||\mathbf{e} - \enc(\mathbf{x_0})||_2^2$.
This attack can be further improved when some prior of the input is known~\cite{tv, dip}. For instance, images tend to consist mostly of low-frequency components and the optimization problem can be regularized with total variation (TV) prior to reduce high-frequency components in $\hat{\mathbf{x}}$~\cite{tv}.
Alternatively, if samples from the underlying input distribution can be obtained, a DNN that generates $\hat{\mathbf{x}}$ from $\mathbf{e}$ can be trained~\cite{tiger, attacks, blackbox}.


\paragraph{Privacy metrics for instance encoding}
To determine whether an encoding is invertible, the vast majority of prior works simply ran a limited set of attacks and observed the result~\cite{privynet, ressfl}. Such approaches are unreliable as more well-designed future attacks may successfully invert the encoding, even if the set of tested attacks failed~\cite{instahide_broken, neuracrypt_broken}.
Others proposed heuristical privacy metrics without rigorous theoretical arguments, such as distance correlation~\cite{nopeek, nopeek-infer} or mutual information~\cite{shredder} between the input and the encoding. While these metrics intuitively make sense, these works failed to show how these metrics are theoretically related to any concrete definition of privacy.

Given these limitations, it is of both interest and practical importance to propose privacy metrics that can theoretically bound the invertibility of instance encoding.
Differential privacy~\cite{dwork2006calibrating, dpbook}, one of the most popular frameworks to quantify privacy in ML~\cite{dpsgd}, is not suitable for instance encoding as its formulation aims to guarantee the worst-case indistinguishability of the encoding from two different inputs. Such indistinguishability significantly damages the utility of downstream tasks~\cite{instahide_broken}, which we show in Appendix~\ref{app:dp}.

A concurrent unpublished work~\cite{posthoc} aims to ensure \emph{indistinguishability between semantically similar inputs}. The work designs an encoder that first embeds an input to a low-dimensional manifold and then uses metric-DP~\cite{metricdp}---a weaker variant of DP---to ensure that the embeddings within the radius $R$ in the manifold are ($\epsilon$, $\delta$)-DP~\cite{posthoc}. This privacy definition is orthogonal to ours, and is less intuitive as it involves a parameter $R$ or a notion of \emph{closeness in the manifold}, whose meanings are hard to interpret.
Our privacy metric is more intuitive to use as it directly lower-bounds the reconstruction error, and does not involve additional hyperparameters such as $R$.

\subsection{Fisher Information Leakage}
\label{sec:dfil_background}

Fisher information leakage (FIL; \citet{fil, fil_guo}) is a measure of leakage through a privacy-enhancing mechanism. Let $\mathcal{M}$ be a randomized mechanism on data sample $\mathbf{x}$, and let $\mathbf{o} \sim \mathcal{M}(\mathbf{x})$ be its output. Suppose that the log density function $\log p(\mathbf{o};\mathbf{x})$ is differentiable w.r.t. $\mathbf{x}$ and satisfies the following regularity condition:
\begin{equation}
\mathbb{E}_\mathbf{o} \left[ \left. \nabla_{\mathbf{x}} \log p(\mathbf{o};\mathbf{x}) \right| \mathbf{x} \right] = 0.
\label{eq:fil_regularity}
\end{equation}
Then, the \emph{Fisher information matrix} (FIM) $\mathcal{I}_\mathbf{o}(\mathbf{x})$ is:
\begin{equation}
    \mathcal{I}_\mathbf{o}(\mathbf{x}) = \mathbb{E}_\mathbf{o}[\nabla_{\mathbf{x}} \log p(\mathbf{o};\mathbf{x})\nabla_{\mathbf{x}} \log p(\mathbf{o};\mathbf{x})^\top].
\label{eq:fim}
\end{equation}

\paragraph{Cramér-Rao bound} Fisher information is a compelling privacy metric as it directly relates to the mean squared error (MSE) of a reconstruction adversary through the Cramér-Rao bound~\cite{kay1993fundamentals}. In detail, suppose that $\hat{\mathbf{x}}(\mathbf{o})$ is an \emph{unbiased} estimate (or reconstruction) of $\mathbf{x}$ given the output of the randomized private mechanism $\mathbf{o} \sim \mathcal{M}(\mathbf{x})$. Then:
\begin{equation}
    \mathbb{E}_\mathbf{o}[||\hat{\mathbf{x}}(\mathbf{o}) - \mathbf{x}||_2^2/d] \ge \frac{d}{\trace(\mathcal{I}_\mathbf{o}(\mathbf{x}))},
    \label{eq:bound_ub}
\end{equation}
where $d$ is the dimension of $\mathbf{x}$ and $\trace$ is the trace of a matrix.
\citet{fil_guo} defined a scalar summary of the FIM called \emph{diagonal Fisher information leakage} (dFIL):
\begin{equation}
\dfil(\mathbf{x}) = \trace(\mathcal{I}_\mathbf{o}(\mathbf{x})) / d,
\label{eq:dfil}
\end{equation}
hence the MSE of an unbiased reconstruction attack is lower bounded by the reciprocal of dFIL. Importantly, dFIL varies with the input $\mathbf{x}$, allowing it to reflect the fact that certain samples may be more vulnerable to reconstruction.

\paragraph{Limitations} Although the Cramér-Rao bound gives a mathematically rigorous interpretation of dFIL, it depends crucially on the \emph{unbiasedness} assumption, \emph{i.e.,} $\mathbb{E}_\mathbf{o}[\hat{\mathbf{x}}(\mathbf{o})] = \mathbf{x}$.
In practice, most real-world attacks use either implicit or explicit priors about the data distribution and are \emph{biased} (\emph{e.g.}, attacks using TV prior or a DNN). It is unclear how dFIL should be interpreted in these more realistic settings. In Section \ref{sec:bound_biased}, we give an alternative theoretical interpretation based on the van Trees inequality~\cite{vantrees}, which lower-bounds the MSE of \emph{any} reconstruction adversary.

\section{Quantifying the Invertibility of Encoding}
\label{sec:dfil}

Motivated by the lack of theoretically-principled metrics for measuring privacy, we propose to adapt the Fisher information leakage framework to quantify the privacy leakage of instance encoding.
We show that many existing encoders can be modified minimally to be interpreted with dFIL and the Cramér-Rao bound. Subsequently, we extend the framework by establishing a connection to the classical problem of score estimation, and derive a novel bound for the reconstruction error of \emph{arbitrary} attacks.


\paragraph{Threat model}
We focus on reconstruction attacks that aim to reconstruct the input $\mathbf{x}$ given its encoding $\mathbf{e} = \enc(\mathbf{x})$.
Following the principle of avoiding security by obscurity, we assume that the attacker has full knowledge of the encoder $\enc$ except for the source of randomness. We consider both unbiased attacks and biased attacks that can use arbitrary prior knowledge about the data distribution to reconstruct new samples from the same distribution.

\paragraph{Privacy definition} At a high level, we consider $\enc$ to be private if $\mathbf{x}$ cannot be reconstructed from the encoding $\mathbf{e}$.
While different measures of reconstruction error exist for different domains, we consider the mean squared error (MSE), defined as $||\hat{\mathbf{x}} - \mathbf{x}||_2^2 / d$, as the primary measure. Although MSE does not exactly indicate semantic similarity, it is widely applicable and is often used as a proxy for semantic similarity~\cite{universal_image_quality, word_embedding}.
Preventing low reconstruction MSE does not necessarily protect against other attacks (\emph{e.g.}, property inference~\cite{attacks_luca}), which we leave as future work.

\subsection{Fisher Information Leakage for Instance Encoding}
\label{sec:dfil_encoding}

To adapt the framework of Fisher information to the setting of instance encoding, we consider the encoding function $\enc$ as a privacy-enhancing mechanism (\emph{cf.} $\mathcal{M}$ in Section \ref{sec:dfil_background}) and use dFIL to measure the privacy leakage of the input $\mathbf{x}$ through its encoding $\mathbf{e} = \enc(\mathbf{x})$.
%
%
However, many instance encoders do not meet the regularity conditions in Equation~\ref{eq:fil_regularity}, making dFIL ill-defined. For example, split inference, split learning, vFL, and \citet{neuracrypt, dauntless} all use DNNs as encoders. DNNs do not produce randomized output, and their log density function $\log p(\mathbf{o};\mathbf{x})$ may not be differentiable when operators like ReLU or max pooling are present.  

Fortunately, many popular encoders can meet the required conditions with small changes.
For example, DNN-based encoders can be modified by (1) replacing any non-smooth functions with smooth functions (\emph{e.g.}, $\tanh$ or GELU~\cite{gelu} instead of ReLU, average pooling instead of max pooling), and (2) adding noise at the end of the encoder for randomness.
%
In particular, if we add random Gaussian noise to a deterministic encoder $\enc_{D}$ (\emph{e.g.,} DNN):
$\enc(\mathbf{x})=\enc_{D}(\mathbf{x}) + \mathcal{N}(0, \sigma^2)$, the FIM of the encoder becomes~\cite{fil}:
\begin{equation}
\mathcal{I}_{\mathbf{e}}(\mathbf{x}) = \frac{1}{\sigma^2}\mathbf{J}_{\enc_D}^{\top}(\mathbf{x})\mathbf{J}_{\enc_D}(\mathbf{x}),
\label{eq:fim_gauss}
\end{equation}
where $\mathbf{J}_{\enc_D}$ is the Jacobian of $\enc_D$ with respect to the input $\mathbf{x}$ and can be easily computed using a single backward pass.
Other (continuously) differentiable encoders can be modified similarly.
Then, Equation~\ref{eq:bound_ub} can be used to bound the reconstruction error, \emph{provided the attack is unbiased}.
%



\subsection{Bounding the Reconstruction of Arbitrary Attacks}
\label{sec:bound_biased}

As mentioned in Section \ref{sec:dfil_background}, most realistic reconstruction attacks are biased, and thus their reconstruction MSE is not lower bounded by the Cramér-Rao bound (Equation~\ref{eq:bound_ub}). As a concrete example, consider an attacker who knows the mean $\mu$ of the input data distribution. If the attacker simply outputs $\mu$ as the reconstruction of any input $\mathbf{x}$, the expected MSE will be the variance of the data distribution \emph{regardless of dFIL}.
Cramér-Rao bound is not applicable in this case because $\mu$ is a biased estimate of $\mathbf{x}$ unless $\mathbf{x} = \mu$.
The above example shows a crucial limitation of the Cramér-Rao bound interpretation of dFIL: it does not take into account any \emph{prior information} the adversary has about the data distribution, which is abundant in the real world (Section~\ref{sec:bg_encoders}).

\paragraph{Bayesian interpretation of Fisher information}
The interpretation of dFIL considered in \citet{fil_guo} (Equation~\ref{eq:bound_ub}) relies on the unbiased attacker assumption, which can be unrealistic for real-world attackers that often employ data priors. Here, we adopt a Bayesian interpretation of dFIL as the difference between an attacker's prior and posterior estimate of the input $\mathbf{x}$. This is achieved through the classical \emph{van Trees inequality}~\cite{vantrees}.
We state the van Trees inequality in Appendix~\ref{app:vantrees}, and use it to derive our MSE bound for arbitrary attacks below as a corollary; proof is in Appendix~\ref{app:proof}.

\begin{corollary}
\label{thm:mse_van_trees}
Let $\pi$ be the input data distribution and let $f_{\pi}(\mathbf{x})$ denote the density function of $\pi$ with respect to Lebesgue measure. Suppose that $\pi$ satisfies the regularity conditions of van Trees inequality (Theorem \ref{thm:van_trees}), and let 
\[\mathcal{J}(f_\pi) = \mathbb{E}_{\pi}[\nabla_{\mathbf{x}} \log f_\pi(\mathbf{x})\nabla_{\mathbf{x}} \log f_\pi(\mathbf{x})^{\top}] \]
denote the information theorist's Fisher information~\cite{vantrees_family} of $\pi$. For a private mechanism $\mathcal{M}$ and any reconstruction attack $\hat{\mathbf{x}}(\mathbf{o})$ operating on $\mathbf{o} \sim \mathcal{M}(\mathbf{x})$:
\begin{equation}
\mathbb{E}_{\pi}\mathbb{E}[||\hat{\mathbf{x}} - \mathbf{x}||_2^2/d] \ge \frac{1}{\mathbb{E}_{\pi}[\dfil(\mathbf{x})] + \trace(\mathcal{J}(f_\pi)) / d}.
\label{eq:bound_b}
\end{equation}
\end{corollary}

\paragraph{Implications of Corollary~\ref{thm:mse_van_trees}} We can readily apply Corollary \ref{thm:mse_van_trees} to the use case of instance encoding by replacing $\mathcal{M}$ with $\enc$ and $\mathbf{o}$ with $\mathbf{e}$, as outlined in Section \ref{sec:dfil_encoding}. Doing so leads to several interesting practical implications:

1. Corollary \ref{thm:mse_van_trees} is a population-level bound that takes expectation over $\mathbf{x} \sim \pi$. This is necessary because given any \emph{fixed} sample $\mathbf{x}$, there is always an attack $\hat{\mathbf{x}}(\mathbf{e}) = \mathbf{x}$ that perfectly reconstructs $\mathbf{x}$ without observing the encoding $\mathbf{e}$. Such an attack would fail in expectation over $\mathbf{x} \sim \pi$.

2. The term $\mathcal{J}(f_\pi)$ captures prior knowledge about the input. When $\mathcal{J}(f_\pi)=0$, the attacker has no prior information about $\mathbf{x}$, and Corollary \ref{thm:mse_van_trees} reduces to the unbiased bound in Equation \ref{eq:bound_ub}. When $\mathcal{J}(f_\pi)$ is large, the bound becomes small regardless of $\mathbb{E}_{\pi}[\dfil(\mathbf{x})]$, indicating that the attacker can simply guess with the input prior and achieve a low MSE.

3. dFIL can be interpreted as capturing how much \emph{easier} reconstructing the input becomes after observing the encoding ($\mathbb{E}_{\pi}[\dfil(\mathbf{x})]$ term) as opposed to only having knowledge of the input distribution ($\trace(\mathcal{J}(f_\pi)) / d$ term). 
%

%

\paragraph{Estimating $\mathcal{J}(f_\pi)$}
The term $\mathcal{J}(f_\pi)$ captures the prior knowledge of the input and plays a crucial role in Corollary \ref{thm:mse_van_trees}. In simple cases where $\pi$ is a known distribution whose density function follows a tractable form, (\emph{e.g.}, when the input follows a Gaussian distribution), $\mathcal{J}(f_\pi)$ can be directly calculated. In such settings, Corollary \ref{thm:mse_van_trees} gives a meaningful theoretical lower bound for the reconstruction MSE.
%

However, most real-world data distributions do not have a tractable form and $\mathcal{J}(f_\pi)$ must be estimated from data. Fortunately, the $\nabla_{\mathbf{x}} \log f_\pi(\mathbf{x})$ term in $\mathcal{J}(f_\pi)$ is a well-known quantity called the \emph{score function}, and there exists a class of algorithms known as \emph{score matching}~\cite{hyvarinen2005estimation, li2017gradient, sliced_score_matching} that aim to estimate the score function given samples from the data distribution $\pi$. We leverage these techniques to estimate $\mathcal{J}(f_\pi)$ when it cannot be calculated; details are in Appendix \ref{app:score_matching}.

\paragraph{Using Corollary \ref{thm:mse_van_trees} in practice}
When $\mathcal{J}(f_\pi)$ is known (\emph{e.g.}, Gaussian), the bound from Corollary~\ref{thm:mse_van_trees} always hold.
%
However, when estimating $\mathcal{J}(f_\pi)$ from data, it can \emph{underestimate} the prior knowledge an attacker can have, leading to an
\emph{incorrect} bound. This can happen due to several reasons, including  improper modeling of the score function, violations of the van Trees regularity conditions, or not having enough representative samples.
The bound can also be loose when tightness conditions of the van Trees do not hold.
%

Even when the bound is not exact, however, Equations~\ref{eq:bound_ub} and \ref{eq:bound_b} can still be interpreted to suggest that increasing $1/\dfil$ strictly makes reconstruction harder.
Thus, we argue that dFIL still serves as a useful privacy metric that in theory bounds the invertibility of an instance encoding. When not exact, the bound should be viewed more as a \emph{guideline} for interpreting and setting dFIL in a data-dependent manner.

\subsection{Evaluation of the Bound}
\label{sec:eval_bound}



We show that Corollary~\ref{thm:mse_van_trees} accurately reflects the reconstruction MSE on both (1) synthetic data with known $\mathcal{J}(f_\pi)$, and (2) real world data with estimated $\mathcal{J}(f_\pi)$.

\subsubsection{Synthetic Data with Known $\mathcal{J}(f_\pi)$}
\label{sec:eval_b_known}

\paragraph{Evaluation setup}
We consider a synthetic Gaussian input distribution: $\textbf{x} \sim \mathcal{N}(0, \tau^2\mathbf{I}_d)$ with $d=784$ and $\tau=0.05$. It can be shown that $\trace(\mathcal{J}(f_\pi)) / d = 1 / \tau^2$, hence a larger $\tau$ forces the data to spread out more and reduces the input prior.
We use a simple encoder which randomly projects the data to a $10,000$-dimensional spaces and then adds Gaussian noise, \emph{i.e.}, $\mathbf{e} = \mathbf{M}\mathbf{x} + \mathcal{N}(0, \sigma^2)$, where $\mathbf{M} \in \mathbb{R}^{10,000 \times 784}$.

\paragraph{Attacks} We evaluate our bound against two different attacks. An unbiased attack (\emph{Attack-ub}) solves the following optimization:
$\hat{\mathbf{x}}(\mathbf{e}) = \underset{\mathbf{x_0}}{\arg\min}||\mathbf{e} - \enc(\mathbf{x_0})||_2^2$.
The attack is unbiased as the objective is convex, and $\mathbf{x}$ is recovered in expectation.
%
A more powerful biased attack (\emph{Attack-b}) adds a regularizer term $\lambda \log p_{\tau}(\mathbf{x}_0)$ to the above objective, where $p_{\tau}$ is the density function of $\mathcal{N}(0, \tau^2\mathbf{I}_d)$. One can show that with a suitable choice of $\lambda$, this attack returns the \emph{maximum a posteriori} estimate of $\mathbf{x}$, which leverages knowledge of the input distribution. Details are in Appendix~\ref{app:hyperparams}.

\begin{figure}[t]
    \centering
    \includegraphics[width=.49\textwidth]{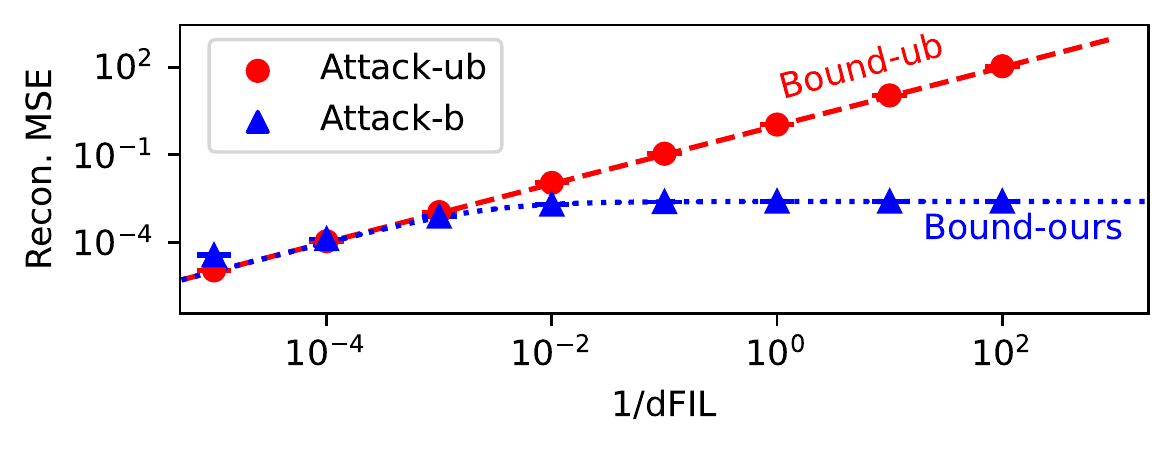}
    \vspace{-4ex}
    \caption{Corollary~\ref{thm:mse_van_trees} holds for synthetic Gaussian dataset, while the bound from prior work only works for unbiased attacks.}
    \label{fig:eval_b_gauss}
    \vspace{-1ex}
\end{figure}

\paragraph{Result}
Figure~\ref{fig:eval_b_gauss} plots the MSE of the two attacks, and the bounds for unbiased (Equation \ref{eq:bound_ub}) and arbitrary attack (Equation \ref{eq:bound_b}).
The MSE of \emph{Attack-ub} (red circle) matches the unbiased attack lower bound (\emph{Bound-ub}; red dashed line), showing the predictive power of Equation \ref{eq:bound_ub} against this restricted class of attacks.
Under \emph{Attack-b} (blue triangle), however, \emph{Bound-ub} breaks. 
Our new bound from Equation~\ref{eq:bound_b} (\emph{Bound-ours}, blue dotted line) reliably holds for both attacks, initially being close to the unbiased bound and converging to guessing only with the input prior (attaining $\tau^2$).
%
%
%


\subsubsection{Real World Data with Estimated $\mathcal{J}(f_\pi)$}
\label{sec:eval_b_unknown}

\paragraph{Evaluation setup}
We also evaluated Corollary~\ref{thm:mse_van_trees} on MNIST~\cite{mnist} and CIFAR-10~\cite{cifar10}. Here, we estimated $\mathcal{J}(f_\pi)$ using sliced score matching~\cite{sliced_score_matching}.
As discussed in Appendix~\ref{app:score_matching}, a moderate amount of randomized smoothing (adding Gaussian noise to the raw input; \citet{randomized_smoothing}) is necessary to ensure that the score estimation is stable
and that regularity conditions in van Trees inequality are satisfied.
%
%
We used a simple CNN-based encoder: $\mathbf{e} = \conv(\mathbf{x}) + \mathcal{N}(0, \sigma^2)$.

\begin{figure}[t]
\centering
    \subfigure[1/dFIL vs. reconstruction MSE]{%
    \includegraphics[width=.49\textwidth]{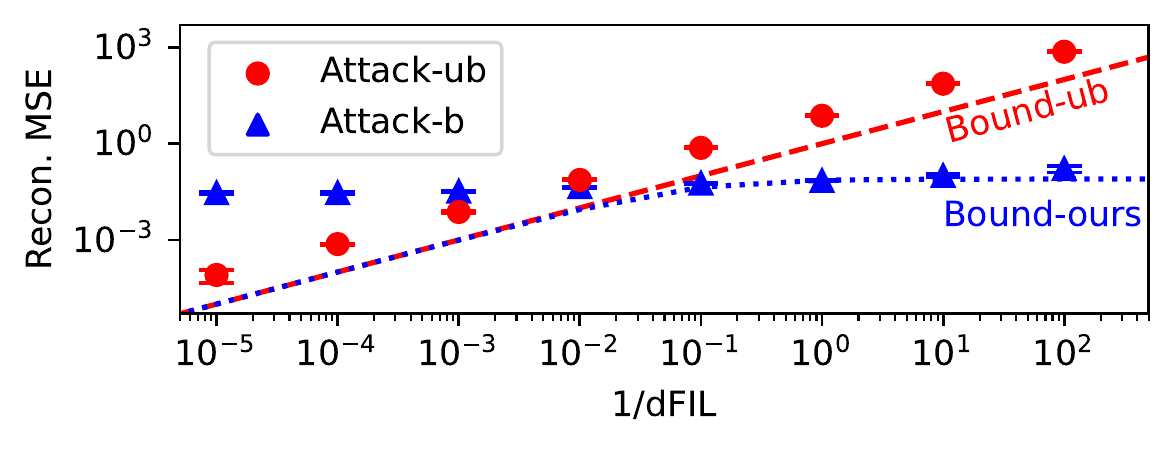}%
    \label{fig:eval_mnist_plt}%
    }
    \subfigure[1/dFIL vs. reconstructed image quality (Attack-biased)]{%
    \includegraphics[width=.49\textwidth]{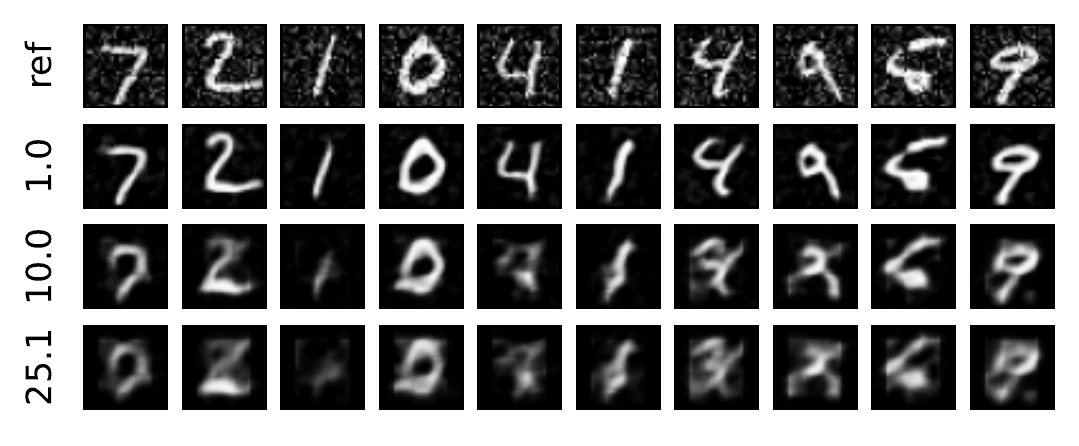}%
    \label{fig:eval_mnist_img}%
    }
    \vspace{-2ex}
    \caption{Corollary~\ref{thm:mse_van_trees} holds for MNIST dataset with a randomized smoothing noise of $\mathcal{N}(0, 0.25^2)$.}
    \label{fig:eval_mnist}
    \vspace{-4ex}
\end{figure}

\paragraph{Attacks} We evaluated \emph{Attack-ub}, which is the same as in Section~\ref{sec:eval_b_known}, and \emph{Attack-b}, which is a trained DNN that outputs the reconstruction given an encoding~\cite{ressfl}. We also evaluated regularization-based attacks~\cite{tv, dip} and obtained similar results; we omit those results for brevity.
%
%

\begin{figure}[t]
\centering
    \subfigure[1/dFIL vs. reconstruction MSE]{%
    \includegraphics[width=.49\textwidth]{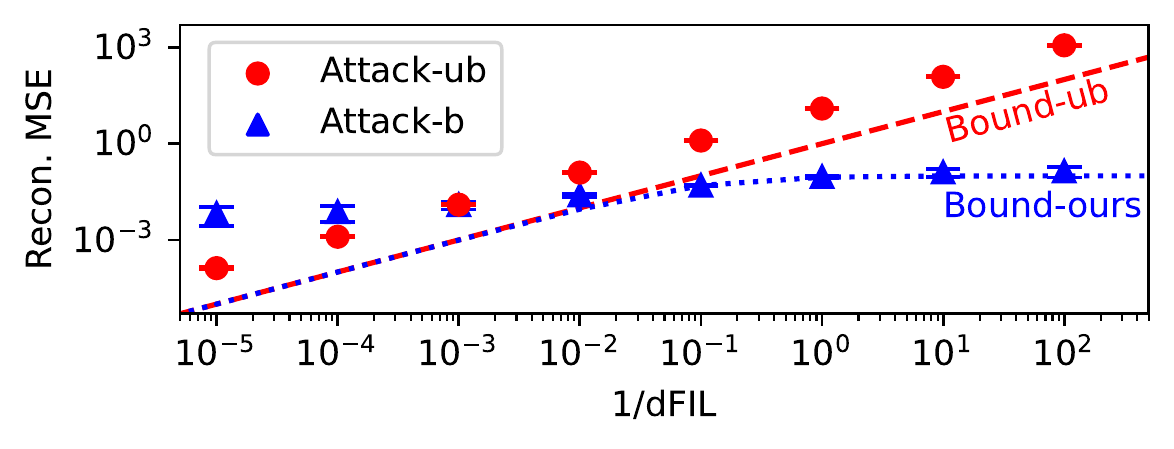}%
    \label{fig:eval_cifar_plt_noise}%
    }
    \subfigure[1/dFIL vs. reconstructed image quality (Attack-biased)]{%
    \includegraphics[width=.49\textwidth]{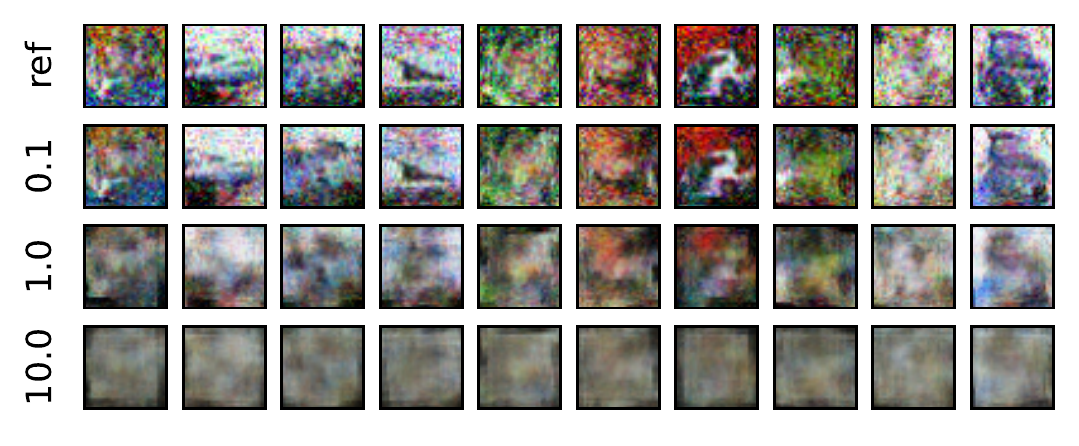}%
    \label{fig:eval_cifar_img_noise}%
    }
    \caption{Corollary~\ref{thm:mse_van_trees} holds for CIFAR-10 dataset with a randomized smoothing noise of $\mathcal{N}(0, 0.25^2)$.}
    \label{fig:eval_cifar_noise}
\end{figure}

\paragraph{Result}
Figures~\ref{fig:eval_mnist_plt} and \ref{fig:eval_cifar_plt_noise} plot the result with a randomized smoothing noise of $\mathcal{N}(0, 0.25^2)$.
Again, \emph{Bound-ub} correctly bounds the MSE achieved by \emph{Attack-ub}. While \emph{Attack-b} is not as effective for very low $1/\dfil$, it outperforms \emph{Attack-ub} for high $1/\dfil$, breaking \emph{Bound-ub}. In comparison, Corollary \ref{thm:mse_van_trees} estimated using score matching (\emph{Bound-ours}) gives a valid lower bound for both attacks.
%
%

Figures~\ref{fig:eval_mnist_img} and \ref{fig:eval_cifar_img_noise} highlights some of the reconstructions visually. Here, the left-hand side number indicates the target $1/\dfil$ and the images are reconstructed using \emph{Attack-b}. In both figures, it can be seen that dFIL correlates well with the visual quality of reconstructed images, with higher values of $1/\dfil$ indicating less faithful reconstructions. See Appendix: Figure~\ref{fig:eval_mnist_full}--\ref{fig:eval_cifar_full} for more results.

\begin{figure}[t]
\centering
    \subfigure[1/dFIL vs. reconstruction MSE]{%
    \includegraphics[width=.49\textwidth]{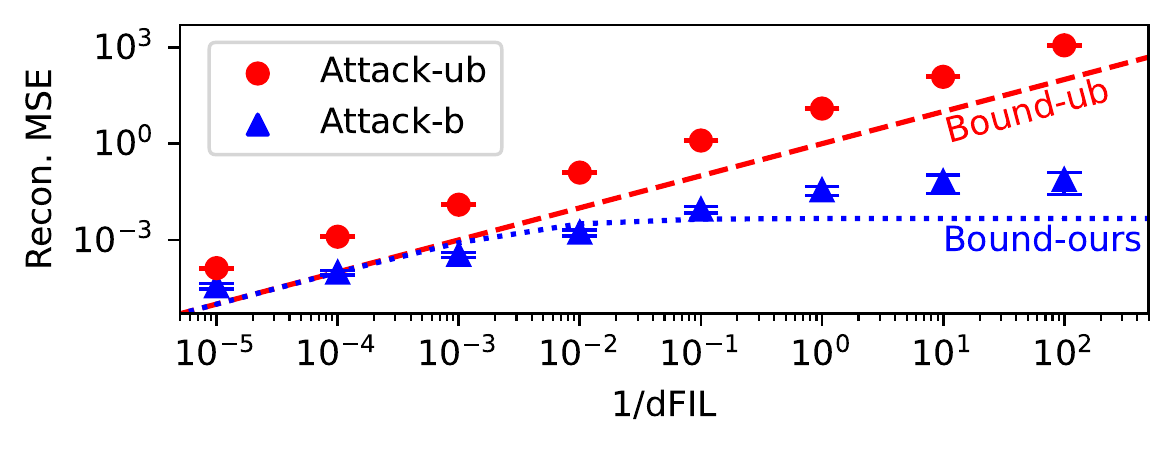}%
    \label{fig:eval_cifar_plt_fail}%
    }
    \subfigure[1/dFIL vs. reconstructed image quality (Attack-biased)]{%
    \includegraphics[width=.49\textwidth]{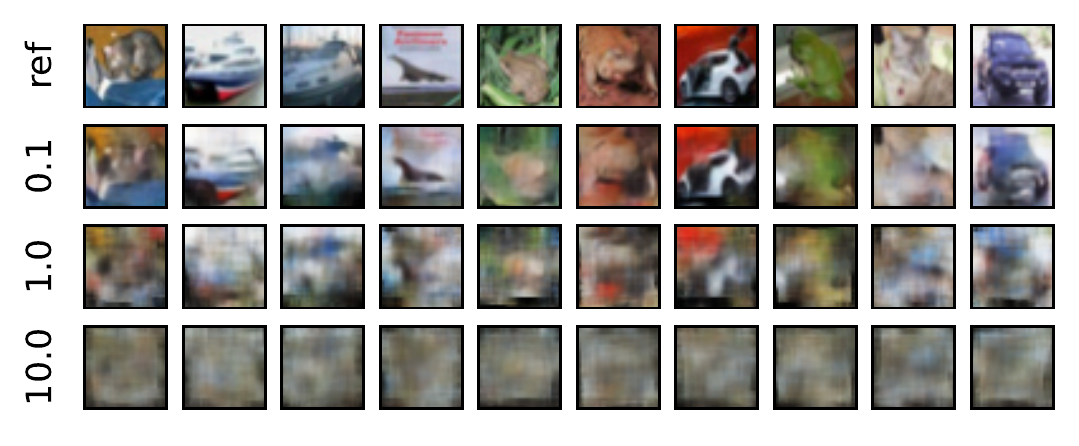}%
    \label{fig:eval_cifar_img_fail}%
    }
    \caption{Corollary~\ref{thm:mse_van_trees} breaks for CIFAR-10 dataset with a randomized smoothing noise of $\mathcal{N}(0, 0.01^2)$. Nonetheless, dFIL shows a strong correlation with the reconstruction quality.}
    \vspace{-15pt}
    \label{fig:eval_cifar_fail}
\end{figure}

Figure~\ref{fig:eval_cifar_fail} additionally shows the result with a much smaller randomized smoothing noise of $\mathcal{N}(0, 0.01^2)$. Unlike previous results, \emph{Bound-ours} breaks around $1/\dfil$=$10^{-3}$.
We suspect it is due to score matching failing when the data lie on a low-dimensional manifold and the likelihood changes rapidly near the manifold boundary, which can be the case when the smoothing noise is small.
The bound is also looser near $1/\dfil$=$10^2$.
Nonetheless, the bound still correlates well with actual attack MSE and the visual reconstruction quality. For these reasons, we claim that dFIL still serves as a useful privacy metric, with a theoretically-principled interpretation and a strong empirical correlation to invertibility.
More reconstructions are shown in Appendix: Figure~\ref{fig:eval_cifar_0_01_full}.




\section{Case Study 1: Split Inference with dFIL}
\label{sec:split_inference}

In the following sections, we discuss two concrete use cases of instance encoding: split inference and training on encoded data. We measure and control privacy using dFIL, and show that it gives useful privacy semantics in practice.

\subsection{Private Split Inference with dFIL}
\label{sec:split_inference_sys}

Split inference~\cite{neurosurgeon, autosplit, nopeek-infer} is a method to run inference of a large DNN that is hosted on the server, without the client disclosing raw input. It is done by running the first few layers of a large DNN on the client device and sending the intermediate activation, instead of raw data, to the server to complete the inference.
The client computation can be viewed as instance encoding, where the first few layers on the client device act as an encoder. However, without additional intervention, split inference by itself is \emph{not} private because the encoding can be inverted~\cite{attacks}.

We design a private split inference system by measuring and controlling the invertibility of the encoder with dFIL.
%
Because the encoder of split inference is a DNN, dFIL can be calculated using Equation~\ref{eq:fim_gauss} with minor modifications to the network (see Section \ref{sec:dfil_encoding}), and can be easily controlled by adjusting the amount of added noise.

\paragraph{Optimizations.} There are several optimizations that can improve the model accuracy for the same dFIL.

1. We calculate the amount of noise that needs to be added to the encoding to achieve a target dFIL, and add a similar amount of noise during training.

2. For CNNs, we add a compression layer---a convolution layer that reduces the channel dimension significantly---at the end of the encoder and a corresponding decompression layer at the beginning of the server-side model. Similar heuristics were explored in \citet{splitnets, ressfl} to reduce the encoder's information leakage.

3. We add an \emph{SNR regularizer} that is designed to maximize the signal-to-noise ratio of the encoding.
From Equations~\ref{eq:dfil}--\ref{eq:fim_gauss}, the noise that needs to be added to achieve a certain dFIL is  $\sigma=\sqrt{\trace(\mathbf{J}_{\enc_D}^{\top}(\mathbf{x})\mathbf{J}_{\enc_D}(\mathbf{x}))/(d * \dfil)}$.
Thus, maximizing the signal-to-noise ratio (SNR) of the encoding ($\mathbf{e}^{\top} \mathbf{e} / \sigma^2$) is equivalent to minimizing 
    $\frac{\trace(\mathbf{J}_{\enc_D}^{\top}(\mathbf{x})\mathbf{J}_{\enc_D}(\mathbf{x}))}{\mathbf{e}^{\top} \mathbf{e}}$,
which we add to the optimizer during training.
%
%

These optimizations were selected from comparing multiple heuristics from prior work~\cite{noise1, noise2, privynet, nopeek-infer, ressfl, splitnets}, and result in a notable reduction of dFIL for the same level of test accuracy.


\subsection{Evaluation of dFIL-based Split Inference}
\label{sec:split_inference_eval}

%
We evaluate our dFIL-based split inference systems' empirical privacy (Section~\ref{sec:eval_split_inference_privacy}) and utility (Section~\ref{sec:eval_split_inference_accuracy}).

\subsubsection{Evaluation Setup}
\label{sec:eval_split_inference_setup}

\begin{figure}[t]
\centering
    \subfigure[1/dFIL vs. SSIM (higher means successful reconstruction)]{%
    \includegraphics[width=.49\textwidth]{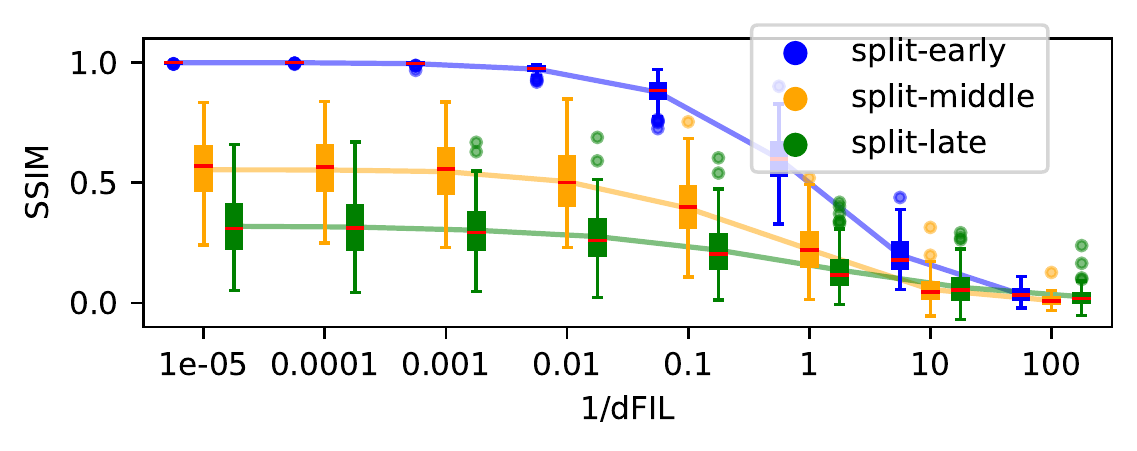}%
    \label{fig:eval_cifar_plt_more}%
    }
    \subfigure[1/dFIL vs. reconstructed image quality]{%
    \includegraphics[width=.49\textwidth]{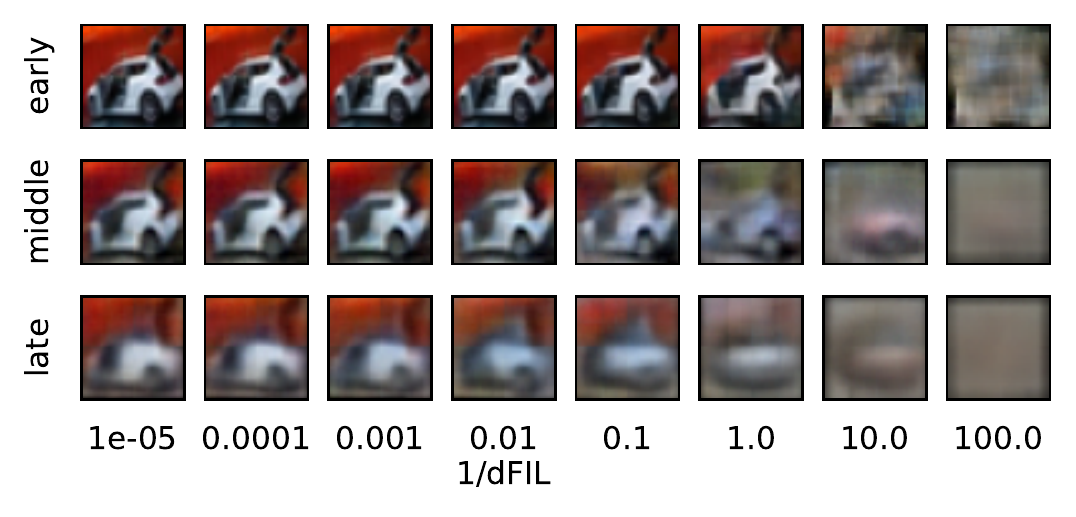}%
    \label{fig:eval_cifar10_img_more}%
    }
    \caption{dFIL and the reconstruction image quality have a strong correlation (1) for metrics other than MSE and (2) qualitatively.}
    \label{fig:eval_cifar_more}
\end{figure}

\paragraph{Models and datasets}
We used three different models and datasets to cover a wide range of applications: ResNet-18~\cite{resnet} with CIFAR-10~\cite{cifar10} for image classification, MLP-based neural collaborative filtering (NCF-MLP)~\cite{neumf} with MovieLens-20M~\cite{movielens} for recommendation, and DistilBert~\cite{distilbert} with GLUE-SST2~\cite{glue_sst2} for sentiment analysis.
See Appendix~\ref{app:hyperparams} for details.

\paragraph{Detailed setups and attacks}

For ResNet-18, we explored three split inference configurations: splitting early (after the first convolution layer), in the middle (after block 4), and late (after block 6). 
We evaluated the empirical privacy with a DNN attacker~\cite{ressfl} and measured the reconstruction quality with structural similarity index measure (SSIM)~\cite{ssim}.
Other popular attacks showed similar trends (see Appendix: Figure~\ref{fig:eval_cifar_more_tv}).

NCF-MLP translates a user id (uid) and a movie id (mid) into embeddings with an embedding table and sends them through a DNN to make a prediction.
We split the NCF-MLP model after the first linear layer of the MLP and tried reconstructing the original uid and mid from the encoding. This is done by first reconstructing the embeddings from the encoding using direct optimization ($\hat{{emb}}(\mathbf{e}) = \underset{{emb_0}}{\arg\min}||\mathbf{e} - \enc({emb_0})||_2^2$), and finding the original uid and mid by finding the closest embedding value in the embedding table: $id = \underset{i}{\arg\min}||\hat{emb} - Emb[i]||_2^2$, where $Emb[i]$ is the $i$-th entry of the embedding table.

For DistilBert, we again explored three different splitting configurations: splitting early (right after block 0), in the middle (after block 2), and late (after block 4).
We use a similar attack to NCF-NLP to retrieve each word token.

\begin{figure*}[t]
    \subfigure[NCF-MLP]{%
    \includegraphics[width=.49\textwidth]{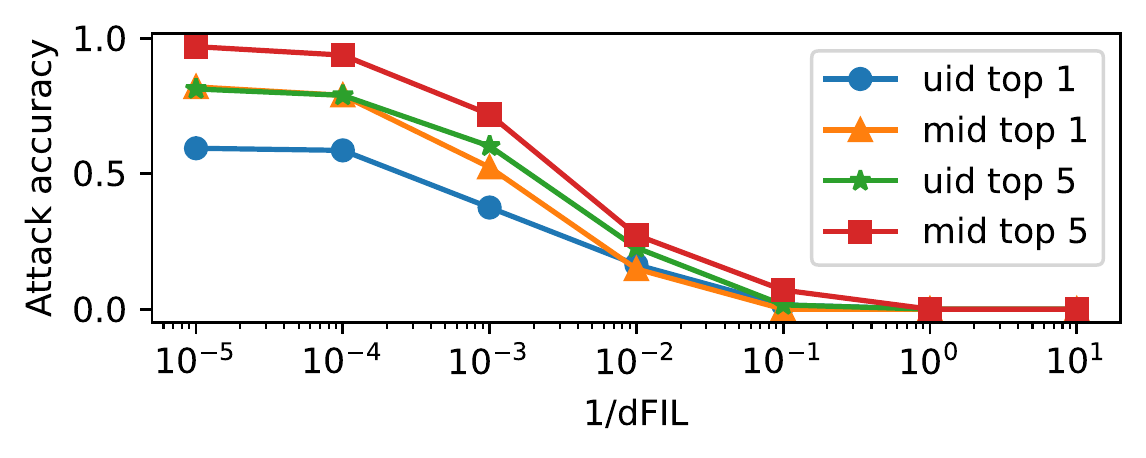}%
    \label{fig:eval_movielens}%
    }
    \subfigure[DistilBert]{%
    \includegraphics[width=.49\textwidth]{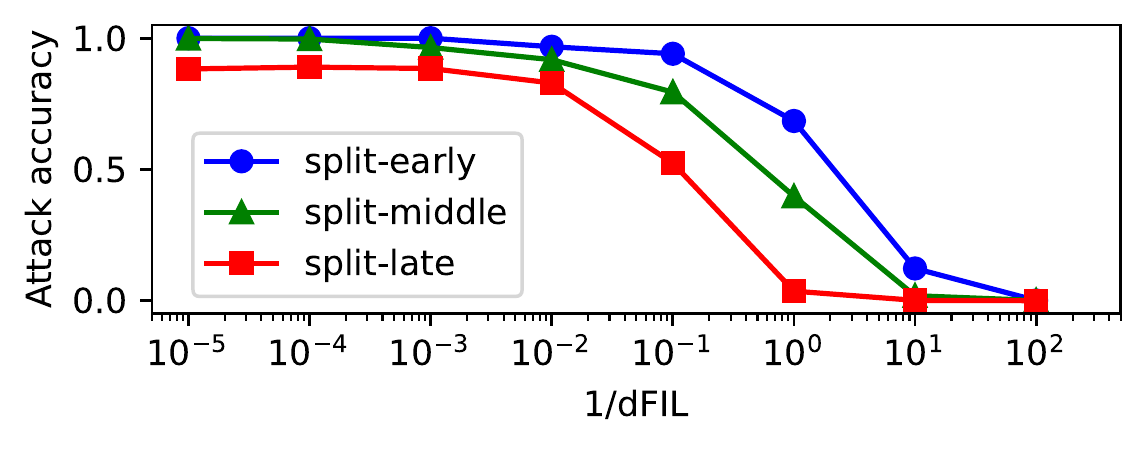}%
    \label{fig:eval_transformer}%
    }
    \caption{Attack accuracy vs. $1/\dfil$ for split inference on encoded data. Increasing $1/\dfil$ reduces the attack success rate.}
    \label{fig:eval_ml_sst2}
\end{figure*}

\subsubsection{Privacy Evaluation Results}
\label{sec:eval_split_inference_privacy}

Figure~\ref{fig:eval_cifar_more} shows the attack result for ResNet-18.
Setups with lower dFIL lead to lower SSIM and less identifiable images, indicating that dFIL strongly correlates with the attack success rate.
The figures also show that the privacy leakage estimated by dFIL can sometimes be conservative. Some setups show empirically-high privacy even when dFIL indicates otherwise, especially when splitting late.

Figures~\ref{fig:eval_movielens} and \ref{fig:eval_transformer} show the attack result for NCF-MLP and DistilBert, respectively. Setups with lower dFIL again consistently showed a worse attack success rate.
%
A sample reconstruction for DistilBert is shown in Appendix: Table~\ref{tab:recon_texts}.

\subsubsection{Utility Evaluation Result}
\label{sec:eval_split_inference_accuracy}

Table~\ref{tab:split_inference} summarizes the test accuracy of the split inference models, where $1/\dfil$ is chosen so that the attacker's reconstruction error is relatively high.
For the same value of $1/\dfil$, our proposed optimizations (\textbf{Ours} column) improve the accuracy significantly compared to simply adding noise (\textbf{No opt.} column). In general, reasonable accuracy can be achieved with encoders with relatively low dFIL.

Accuracy degrades more when splitting earlier, indicating that more noise is added to the encoding. Prior works showed that splitting earlier makes the reconstruction easier because the encoding is leakier~\cite{tv}. The result indicates that our dFIL-based split inference adds more noise to leakier encodings, as expected.

\begin{table}[]
    \small
    \centering 
    \caption{Test accuracy for different split inference setups with different dFIL. Base accuracy of each model is in the parenthesis.}
    \label{tab:split_inference}
    \begin{tabular}{c|c|c|c|c}
         Setup & Split & $\frac{1}{dFIL}$ & No opt. & Ours \\\hline\hline
         \multirow{6}{*}{\makecell{CIFAR-10\\+ ResNet-18\\(acc: 92.70\%)}} & \multirow{2}{*}{early} & 10 & 10.70\% & \textbf{74.44\%}\\
         & & 100 & 10.14\%  & \textbf{57.97\%}\\\cline{2-5}
         
          & \multirow{2}{*}{middle} & 10 & 22.11\% & \textbf{91.35\%}\\
         & & 100 & 12.94\% & \textbf{84.27\%}\\\cline{2-5}
         
         & \multirow{2}{*}{late} & 10 & 78.48\% & \textbf{92.35\%}\\
         & & 100 & 33.54\% & \textbf{87.58\%}\\\hline
         
         \multirow{3}{*}{\makecell{MovieLens-20M\\+ NCF-MLP\\(AUC: 0.8228)}} & \multirow{3}{*}{early} & 1 & 0.8172 &  \textbf{0.8286}\\
         & & 10 & 0.7459 & \textbf{0.8251}\\
         & & 100 & 0.6120 & \textbf{0.8081}\\\hline

         \multirow{6}{*}{\makecell{GLUE-SST2\\+ DistilBert\\(acc: 91.04\%)}} & \multirow{2}{*}{early} & 10 & 50.80\% & \textbf{82.80\%}\\
         & & 100 & 49.08\% & \textbf{81.88\%} \\\cline{2-5}

          & \multirow{2}{*}{middle} & 10 & 76.61\% & \textbf{83.03\%}\\
         & & 100 & 61.93\% & \textbf{82.22\%}\\\cline{2-5}

         & \multirow{2}{*}{late} & 10 & \textbf{90.25\%} & {83.03\%}\\
         & & 100 & {82.68\%} & \textbf{82.82\%}\\
    \end{tabular}
    \vspace{-2ex}
\end{table}

\section{Case Study 2: Training with dFIL}
\label{sec:private_training}

As a second use case, we consider training a model on privately encoded data with its privacy controlled by dFIL.

\subsection{Training on Encoded Data with dFIL}
\label{sec:private_training_sys}


We consider a scenario where users publish their encoded private data, and a downstream model is trained on the encoded data.
We use the first few layers of a pretrained model as the encoder by freezing the weights and applying the necessary changes in Section \ref{sec:dfil_encoding}. Then, we use the rest of the model with its last layer modified for the downstream task and finetune it with the encoded data.
We found that similar optimizations from split inference (\emph{e.g.,} compression layer, SNR regularizer) benefit this use case as well.


\subsection{Evaluation of dFIL-based Training}
\label{sec:private_training_eval}

We evaluate the model utility and show that it can reach a reasonable accuracy when trained on encoded data. We omit the privacy evaluation as it is similar to Section~\ref{sec:eval_split_inference_privacy}.

\subsubsection{Evaluation Setup}
\label{sec:eval_private_training_setup}

We train a ResNet-18 model for CIFAR-10 classification. The model is pretrained on one of two different datasets: (1) CIFAR-100~\cite{cifar10}, and (2) held-out 20\% of CIFAR-10. Then, layers up to block 4 are frozen and used as the encoder. 
The CIFAR-10 training set is encoded using the encoder and used to finetune the rest of the model.
The setup mimics a scenario where some publicly-available data whose distribution is similar (CIFAR-100) or the same (held-out CIFAR-10) with the target data is available and is used for encoder training.
Detailed hyperparameters are in Appendix~\ref{app:hyperparams}.

\subsubsection{Utility Evaluation Result}
\label{sec:eval_private_training_accuracy}

Table~\ref{tab:private_training} summarizes the result. Our design was able to achieve a decent accuracy using encoded data with relatively safe dFIL values (10--100). The result indicates that model training with privately encoded data is possible.
The achieved accuracy was higher when the encoder was trained with data whose distribution is more similar to the downstream task (CIFAR-10).
We believe more studies in hyperparameter/architecture search will improve the result.

\begin{table}[]
    \small
    \centering 
    \caption{Accuracy from training with different encoders.}
    \label{tab:private_training}
    \begin{tabular}{c|c|c}
         Pretrain dataset & 1/dFIL & Acc. \\\hline\hline
          \multirow{2}{*}{CIFAR-100} & 10 & 80.16\% \\
          & 100 & 70.27\% \\\hline
          \multirow{2}{*}{\makecell{CIFAR-10\\(held-out 20\%)}} & 10 & 81.99\% \\
          & 100 & 78.65\% \\
         
    \end{tabular}
\end{table}
\section{Discussion}

We propose dFIL as a theoretically-principled privacy metric for instance encoding. We show that dFIL can provide a general reconstruction error bound against arbitrary attackers. We subsequently show that training/inference is possible with data privately encoded with low dFIL.

%

\paragraph{Limitations} dFIL has several potential limitations:

1. Corollary~\ref{thm:mse_van_trees} only bounds the  MSE, which might not always correlate well with the semantic quality of the reconstruction. To address this, van Trees inequality can be extended to an absolutely continuous function $\psi(\mathbf{x})$ to bound $\mathbb{E}[||\psi(\hat{\mathbf{x}}) - \psi(\mathbf{x})||_2^2/d]$~\cite{vantrees_application}, which may be used to extend to metrics other than MSE. 

2. Equation~\ref{eq:bound_b} provides an average bound across the input distribution, so MSE may be below the bound for some samples. This is a fundamental limitation of the Bayesian bound (Section~\ref{sec:bound_biased}).
One can dynamically calculate dFIL for each sample and detect/handle such leaky inputs.

3. 
%
For data types where MSE is not directly meaningful or the bound is inaccurate, it may not be straightforward to interpret the privacy of an encoding given its dFIL.
In such cases, acceptable values of dFIL should be determined for each application through further research.
The situation is similar to DP, where it is often not straightforward what privacy parameters (e.g., $\epsilon$, $\delta$) need to be used for privacy~\cite{jayaraman2019evaluating}.
%

4. Systems with the same dFIL may actually have different privacy levels, as the bound from dFIL may be conservative.
Comparing the privacy of two different systems using dFIL should be done with caution because dFIL is a lower bound rather than an accurate privacy measure. 


%

\bibliography{bib}
\bibliographystyle{icml2022}

\newpage
\appendix
\onecolumn
\section{Appendix}

\subsection{Score Matching Details}
\label{app:score_matching}

We found that using score matching~\cite{sliced_score_matching} does not work reliably when the data's structure lies on a low-dimensional manifold (e.g., natural images). We found that applying randomized smoothing~\cite{randomized_smoothing}, which adds Gaussian noise to the image for robust training, helps stabilize score matching as it smoothens the density function. Randomized smoothing also makes the bound tighter.
We observed that adding a reasonable amount of noise (\emph{e.g.,} standard deviation of 0.25, which was originally used in \citet{randomized_smoothing}) works well in general, but adding only small noise (standard deviation of 0.01) does not. We show both results in Section~\ref{sec:eval_b_known}.

\subsection{Hyperparameters}
\label{app:hyperparams}

\paragraph{Attacks}
For attacks in Section~\ref{sec:eval_b_known}, \ref{sec:eval_b_unknown}, and \ref{sec:split_inference_eval}, we used the following hyperparameters.
For the optimizer-based attack for Gaussian synthetic input, we used Adam with lr=$10^{-3}$, and $\lambda$=0.1--100 for the regularizer.
For the optimizer-based attack for NCF-MLP and DistilBert, we used Adam with lr=0.1.
For the DNN-based attack for MNIST and CIFAR-10 (Figure~\ref{fig:eval_mnist}, \ref{fig:eval_cifar_noise}, \ref{fig:eval_cifar_more}), we used a modified DNN from \citet{ressfl}, which uses a series of convolution (Conv) and convolution transpose (ConvT) layers interspersed with leaky ReLU of slope 0.2. All the models were trained for 100 epochs using Adam with lr=$10^{-3}$. Below summarizes the architecture parameters. For DNN-based attacks in Section~\ref{sec:eval_b_unknown}, we put a sigmoid at the end. For the attack in Section~\ref{sec:split_inference_eval}, we do not.

\begin{table}[h]
    \small
    \centering 
    \caption{DNN attacker architectures used in the paper. Output channel dimension ($c_{out}$), kernel size (k), stride (s), and output padding (op) are specified. Input padding was 1 for all layers.}
    \label{tab:attack_models}
    \begin{tabular}{c|c}
         Dataset + encoder & Architecture\\\hline
         MNIST + Conv & 3$\times$Conv($c_{out}$=16, k=3, s=1) + ConvT($c_{out}$=32, k=3, s=1, op=0) + ConvT($c_{out}$=1, k=3, s=1, op=0) \\
         CIFAR-10 + split-early & 3$\times$Conv($c_{out}$=64, k=3, s=1) + ConvT($c_{out}$=128, k=3, s=1, op=0) + ConvT($c_{out}$=3, k=3, s=1, op=0) \\
         CIFAR-10 + split-middle & 3$\times$Conv($c_{out}$=128, k=3, s=1) + ConvT($c_{out}$=128, k=3, s=2, op=1) + ConvT($c_{out}$=3, k=3, s=2, op=1) \\
         CIFAR-10 + split-late & 3$\times$Conv($c_{out}$=256, k=3, s=1) + 2$\times$ConvT($c_{out}$=256, k=3, s=2, op=1) + ConvT($c_{out}$=3, k=3, s=2, op=1) \\
    \end{tabular}
    
\end{table}

\paragraph{Split inference}

Below are the hyperparameters for the models used in Section~\ref{sec:split_inference_eval}.
For ResNet-18, we used an implementation tuned for CIFAR-10 dataset from~\citet{pytorch_cifar10}, with ReLU replaced with GELU and max pooling replaced with average pooling. We used the default hyperparameters from the repository except for the following:  bs=128, lr=0.1, and weight\_decay=$5\times 10^{-4}$. For NCF-MLP, we used an embedding dimension of 32 and MLP layers of output size [64, 32, 16, 1]. We trained NCF-MLP with Nesterov SGD with momentum=0.9, lr=0.1, and batch size of 128 for a single epoch. We assumed 5-star ratings as click and others as non-click. For DistilBert, we used Adam optimizer with a batch size of 16, lr=$2\times10^{-5}$, $\beta_1$=0.9, $\beta_2$=0.999, and $\epsilon=10^{-8}$.
We swept the compression layer channel dimension among {2, 4, 8, 16}, and the SNR regularizer $\lambda$ between $10^{-3}$ and 100. 

\paragraph{Training}
Below are the hyperparameters for the models evaluated in Section~\ref{sec:private_training_eval}. We used the same model and hyperparameters with split inference for training the encoder with the pretraining dataset. Then, we freeze the layers up to block 4 and trained the rest for 10 epochs with CIFAR-10, with lr=$10^{-3}$ and keeping other hyperparameters the same.

\subsection{van Trees Inequality}
\label{app:vantrees}

Below, we restate the van Trees Inequality from~\cite{vantrees_application}, which we use to prove Theorem~\ref{eq:bound_b}.

\begin{theorem}[Multivariate van Trees inequality]
\label{thm:van_trees}

Let $(\mathcal{X}, \mathcal{F}, P_\mathbf{\theta}:\mathbf{\theta} \in \Theta)$ be a family of distributions on a sample space $\mathcal{X}$ dominated by $\mu$.
Let $p(\mathbf{x}|\mathbf{\theta})$ denote the density of $X \sim P_\theta$ and $\mathcal{I}_{\mathbf{x}}(\mathbf{\theta})$ denotes its FIM.
Let $\mathbf{\theta} \in \Theta$ follows a probability distribution $\pi$ with a density $\lambda_\pi(\mathbf{\theta})$ with respect to Lebesgue measure. Suppose that $\lambda_\pi$ and $p(\mathbf{x}|\mathbf{\theta})$ are absolutely $\mu$-almost surely continuous and $\lambda_\pi$ converges to 0 and the endpoints of $\Theta$.
Let $\psi$ be an absolutely continuous function of $\mathbf{\theta}$, and $\psi_n$ an arbitrary estimator of $\psi(\mathbf{\theta})$. Assume regularity conditions from Corollary~\ref{eq:fil_regularity} is met. If we make $n$ observations $\{\mathbf{x_1}, \mathbf{x_2}, ..., \mathbf{x_n}\}$, then:
    
\[ \int_{\Theta} \mathbb{E}_{\mathbf{\theta}}[||\psi_n - \psi(\theta) ||_2^2]\lambda_\pi(\mathbf{\theta}) d\mathbf{\theta} \ge \frac{(\int \divergence \psi(\mathbf{\theta})\lambda_\pi(\mathbf{\theta})d\mathbf{\theta})^2}{n \int \trace(\mathcal{I}_{\mathbf{x}}(\mathbf{\theta}))\lambda_\pi(\mathbf{\theta})d\mathbf{\theta} + \trace(\mathcal{J}(\lambda_\pi))}\]


\label{eq:vantrees}
\end{theorem}

\subsection{Proof of Corollary~\ref{thm:mse_van_trees}}

\begin{proof}
Let $\psi$ be an identity transformation $\psi(\mathbf{\theta}) = \mathbf{\theta}$. For the setup in Corollary~\ref{thm:mse_van_trees}, $n=1$ and $\divergence(\mathbf{x})=d$, so the multivariate van Trees inequality from Theorem~\ref{eq:vantrees} reduces to:

\[ \mathbb{E}_{\pi}\mathbb{E}_{\theta}[||\hat{\mathbf{x}} - \mathbf{x} ||_2^2/d] \ge \frac{d}{\mathbb{E}_{\pi}[\trace(\mathcal{I}_{\mathbf{e}}(\mathbf{x}))] + \trace(\mathcal{J}(f_\pi))}
= \frac{1}{\mathbb{E}_{\pi}[\dfil(\mathbf{x})] + \trace(\mathcal{J}(f_\pi))/d}
\]

\end{proof}

\label{app:proof}

\subsection{Comparison with Differential Privacy.}
\label{app:dp}

Differential privacy~\cite{dpsgd} is not well-suited for instance encoding, as we discuss in Section~\ref{sec:bg_encoders}. We formulate and compare a DP-based instance encoding and compare it with our dFIL-based instance encoding in a split inference setup (Section~\ref{sec:split_inference}) to show that DP-based instance encoding indeed does not work well.

To formulate DP for instance encoding, we define an adjacent set $\mathcal{D}$ and $\mathcal{D}'$ as two differing inputs. A randomized method $\mathcal{A}$ is ($\alpha$, $\epsilon$)-R\'enyi differentially private (RDP) if $D_{\alpha}(\mathcal{A}(\mathcal{D})||\mathcal{A}(\mathcal{D}')) \le \epsilon$ for $D_{\alpha}(P||Q)=\frac{1}{\alpha - 1} \log{\mathbb{E}_{x \sim Q}[(\frac{P(x)}{Q(x)})^\alpha]}$. As DP provides a different privacy guarantee with dFIL, we use the theorem from \citet{fil_guo} to derive an MSE lower bound using DP's privacy metric for an unbiased attacker.
Assuming a reconstruction attack $\hat{\mathbf{x}} = \att(\mathbf{e})$ that reconstructs $\mathbf{x}$ from the encoding $\mathbf{e} = \enc(\mathbf{x})$, repurposing the theorem from \citet{fil_guo} gives:

\begin{equation}
    \mathbb{E}[||\hat{\mathbf{x}} - \mathbf{x}||_2^2/d] \ge \frac{\Sigma_{i=1}^{d}\diam_{i}(\mathcal{X})^2/4d}{e^\epsilon - 1}
    \label{eq:dp_bound}
\end{equation}

for a (2, $\epsilon$)-RDP $\enc$, where $\mathcal{X}$ is the input data space.
We can construct a (2, $\epsilon$)-RDP encoder $\enc_{RDP}$ from a deterministic encoder $\enc_{D}$ by scaling and clipping the encoding adding Gaussian noise, or $\enc_{RDP} = \enc_{D}(\mathbf{x}) / \max(1, \frac{||\enc_{D}(\mathbf{x})||_2}{C}) + \mathcal{N}(0, \sigma^2)$, similarly to \citet{dpsgd}. The noise to be added is $\sigma = \frac{(2C)^2}{\epsilon}$~\cite{renyi}.
Equation~\ref{eq:dp_bound} for DP is comparable to Equation~\ref{eq:bound_ub} for dFIL, and we use the two equations to compare DP and dFIL parameters.
We use Equation~\ref{eq:bound_ub} because \citet{fil_guo} does not discuss the bound against biased attackers.

We evaluate both encoders for split inference using CIFAR-10 dataset and ResNet-18. We split the model after block 4 (split-middle from Section~\ref{sec:eval_split_inference_setup}) and did not add any optimizations discussed in Section~\ref{sec:split_inference} for simplicity. For the DP-based encoder, we retrain the encoder with scaling and clipping so that the baseline accuracy without noise does not degrade.
We ran both models without standardizing the input, which makes $\diam_{i}(\mathcal{X}) = 1$ for all $i$.

\begin{table}[h]
    \small
    \centering 
    \caption{Test accuracy when targeting the same MSE bound.}
    \label{tab:dp}
    \begin{tabular}{c|c|c|c|c}
         Unbiased MSE bound & 1e-5 & 1e-4 & 1e-3 & 1e-2\\\hline
         dFIL-based & \textbf{93.09\%} & \textbf{93.11\%} & \textbf{92.52\%} & \textbf{87.52\%} \\
         DP-based & 64.64\% & 56.68\% & 46.46\% & 33\% \\
    \end{tabular}
    
\end{table}

Table~\ref{tab:dp} compares the test accuracy achieved when targeting the same MSE bound for an unbiased attacker using dFIL and DP, respectively. The result clearly shows that DP degrades the accuracy much more for similar privacy levels (same unbiased MSE bound), becoming impractical very quickly.
DP suffers from low utility because DP is agnostic with the input and the model, assuming a worst-case input and model weights. Our dFIL-based bound uses the information of the input and model weights in its calculation of the bound and can get a tighter bound.

\subsection{Additional Figures and Tables}

\begin{table}[h]
    \small
    \centering 
    \caption{The reconstruction quality of an input string is highly correlated with dFIL. Correct parts are in bold.}
    \label{tab:recon_texts}
    \begin{tabular}{c|c}
         1/dFIL & Reconstructed text (from split-early)\\\hline\hline
         $10^{-5}$ & \textbf{it's a charming and often affecting journey.} \\\hline
         $1$ & \makecell{\textbf{it's} cones \textbf{charming}ound \\\textbf{often affecting journey} closure} \\\hline
         $10$ & \makecell{grounds yuki cum sign\\recklessound fanuche pm stunt} \\
    \end{tabular}
    
\end{table}

\begin{figure}[h]
\centering
    \subfigure[1/dFIL vs. SSIM (higher means successful reconstruction)]{%
    \includegraphics[width=.49\textwidth]{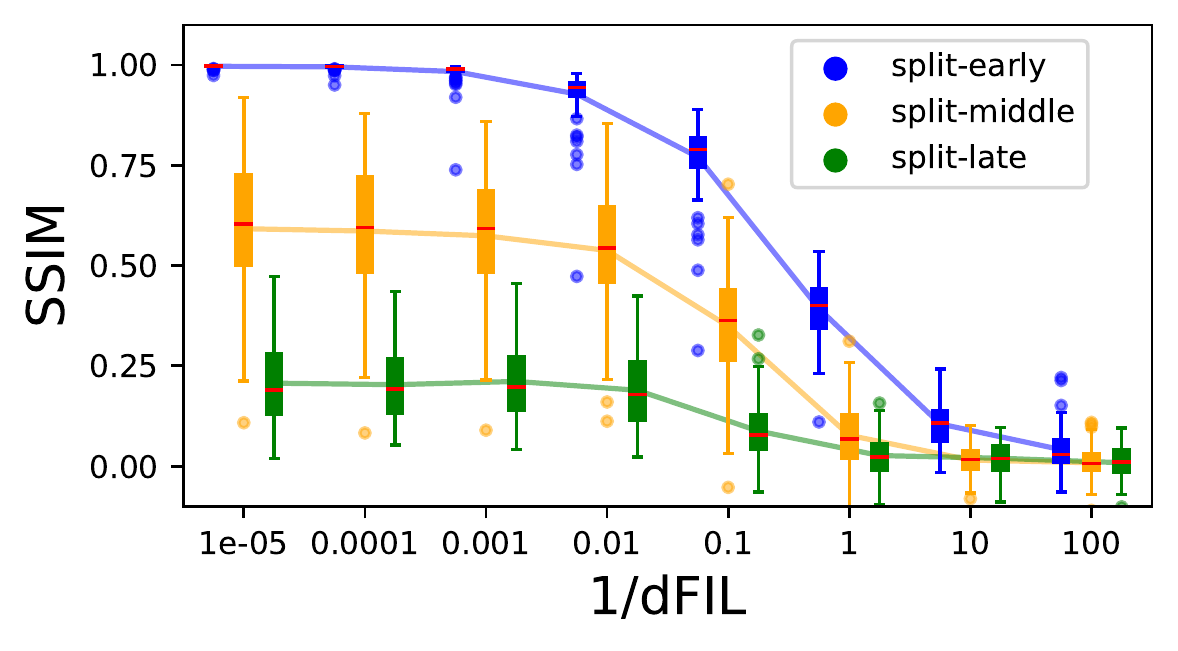}%
    \label{fig:eval_cifar_plt_more_tv}%
    }
    \subfigure[1/dFIL vs. reconstructed image quality]{%
    \includegraphics[width=.49\textwidth]{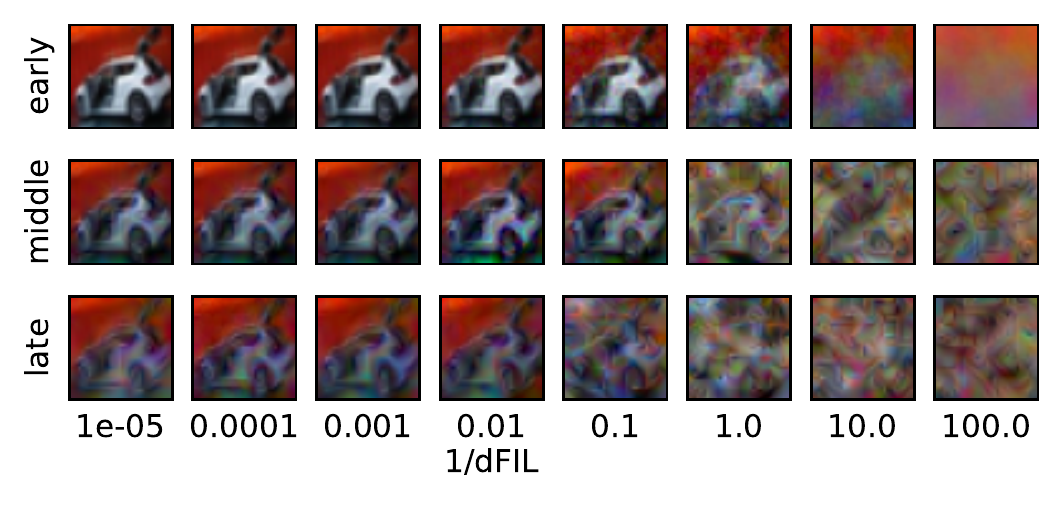}%
    \label{fig:eval_cifar10_img_more_tv}%
    }
    \caption{Optimizer-based attack with total variation (TV) prior~\cite{tv} against our split inference system in Section~\ref{sec:eval_split_inference_privacy}.}
    \label{fig:eval_cifar_more_tv}
\end{figure}

\begin{figure}[t]
    \centering
    \includegraphics[width=.49\textwidth]{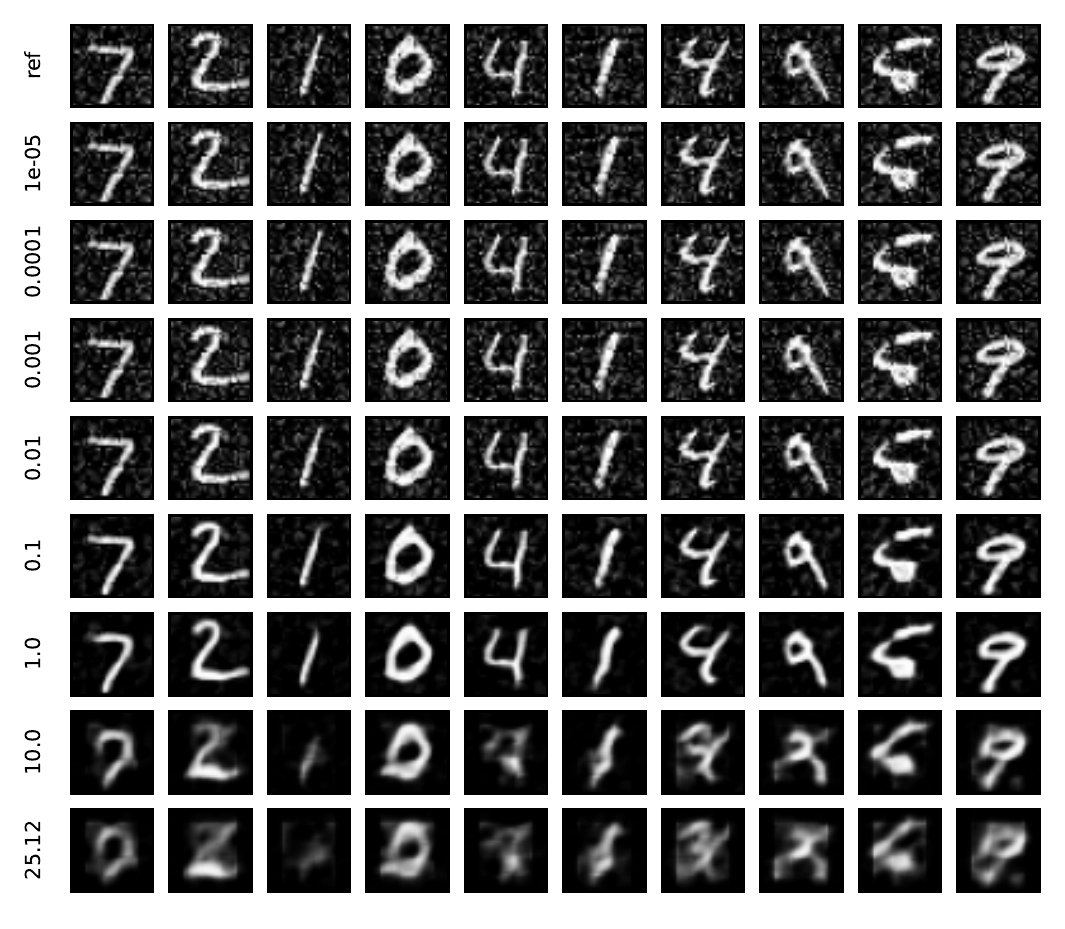}
    \vspace{-4ex}
    \caption{Full reconstruction result of Figure~\ref{fig:eval_mnist_img}.}
    \label{fig:eval_mnist_full}
    \vspace{-1ex}
\end{figure}

\begin{figure}[t]
    \centering
    \includegraphics[width=.49\textwidth]{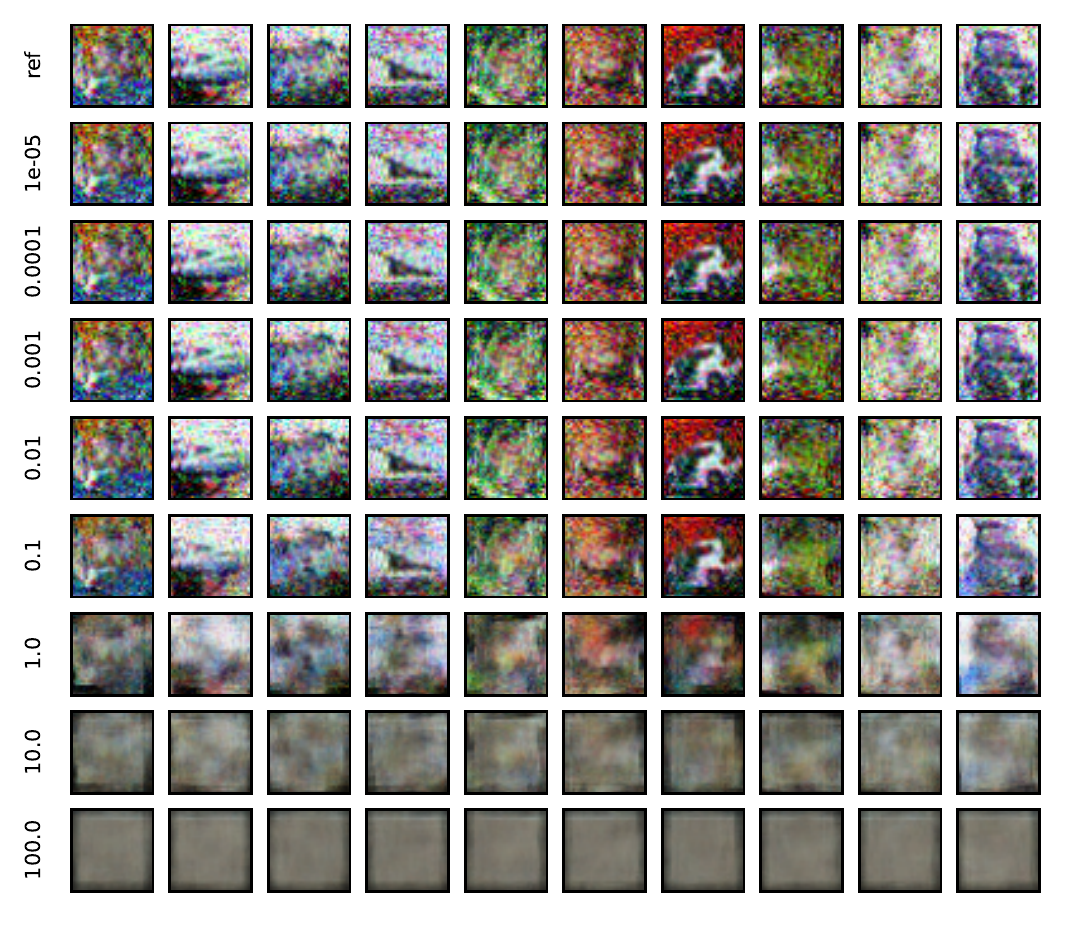}
    \vspace{-4ex}
    \caption{Full reconstruction result of Figure~\ref{fig:eval_cifar_img_noise}.}
    \label{fig:eval_cifar_full}
    \vspace{-1ex}
\end{figure}

\begin{figure}[t]
    \centering
    \includegraphics[width=.49\textwidth]{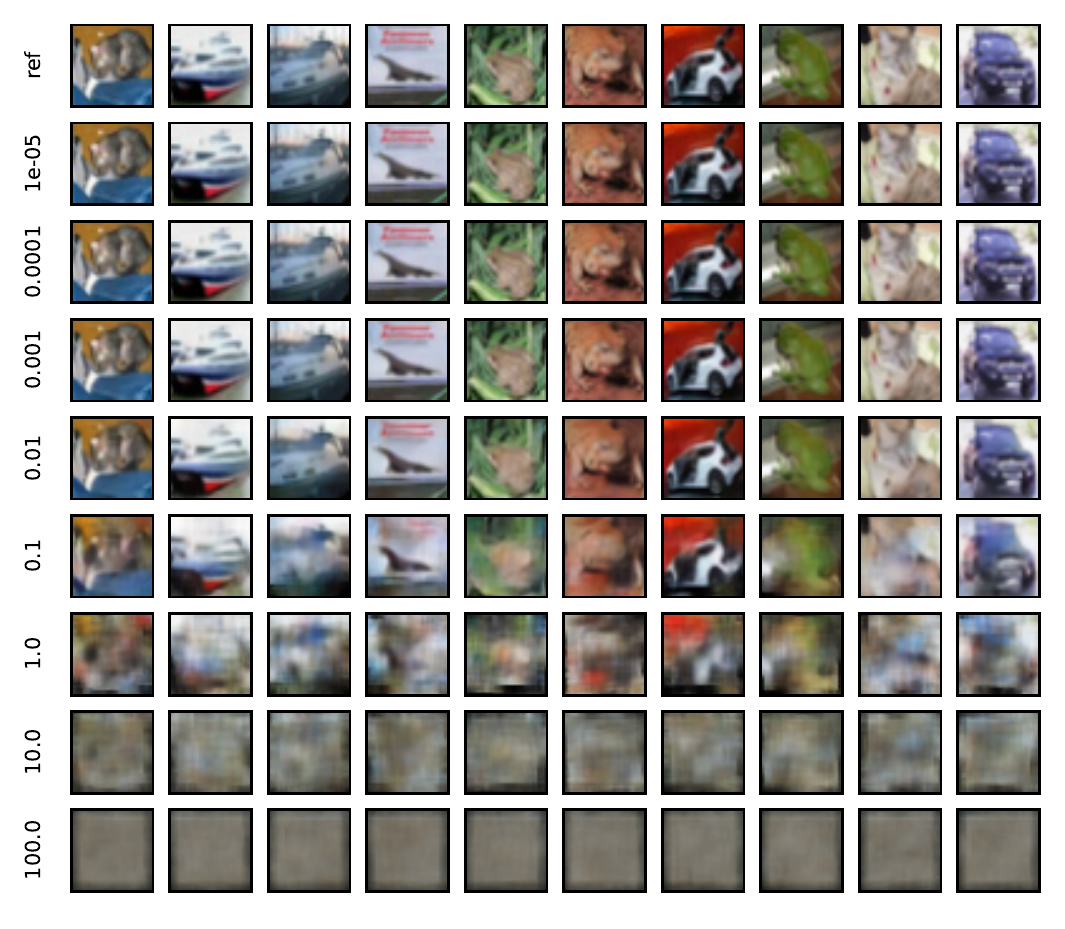}
    \vspace{-4ex}
    \caption{Full reconstruction result of Figure~\ref{fig:eval_cifar_img_fail}.}
    \label{fig:eval_cifar_0_01_full}
    \vspace{-1ex}
\end{figure}

\end{document}